\documentclass[10pt,twocolumn,letterpaper]{article}

\usepackage{iccv}              

\usepackage{algorithm}
\usepackage{algorithmic}

\usepackage{amsthm}  

\theoremstyle{plain}
\newtheorem{theorem}{Theorem}
\newtheorem{proposition}{Proposition}
\newtheorem{lemma}{Lemma}
\newtheorem{corollary}{Corollary}

\theoremstyle{remark}

\definecolor{iccvblue}{rgb}{0.21,0.49,0.74}
\usepackage[pagebackref,breaklinks,colorlinks,allcolors=iccvblue]{hyperref}

\def\paperID{13} 
\def\confName{ICCV}
\def\confYear{2025}

\title{Multi-Agent Pose Uncertainty: A Differentiable Rendering Cram\'{e}r--Rao Bound}

\author{Arun Muthukkumar\\
Illinois Mathematics and Science Academy\\
{\tt\small amuthukkumar@imsa.edu}
}

\begin{document}
\maketitle
\begin{abstract}

Pose estimation is essential for many applications within computer vision and robotics. Despite its uses, few works provide rigorous uncertainty quantification for poses under dense or learned models. We derive a closed-form lower bound on the covariance of camera pose estimates by treating a differentiable renderer as a measurement function. Linearizing image formation with respect to a small pose perturbation on the manifold yields a render-aware Cram\'er--Rao bound. Our approach reduces to classical bundle-adjustment uncertainty, ensuring continuity with vision theory. It also naturally extends to multi-agent settings by fusing Fisher information across cameras. Our statistical formulation has downstream applications for tasks such as cooperative perception and novel view synthesis without requiring explicit keypoint correspondences.

\end{abstract}    
\section{Introduction}

Estimating the 6-DoF pose of a camera from images is foundational for vision and robotics. Neural rendering (NeRF~\cite{mildenhall2020}, Instant-NGP~\cite{mueller2022}, 3D Gaussian Splatting~\cite{kerbl2023}) can offer a dense, differentiable photometric measurement model where each pixel depends on the pose. Works such as iNeRF \cite{Lin2021IROS} found that we may “invert” the renderer to localize cameras by photometric alignment. Despite this rapid progress, there is little theory quantifying pose accuracy from these dense renderers, or how scene content (texture, depth variation, symmetries) fundamentally limits identifiability. To our knowledge, no prior work has derived closed-form \emph{pose CRBs for dense differentiable renderers}.

Classical geometric vision provides a natural lens to answer this question. The Cram\'er--Rao bound (CRB) lower-bounds the covariance of any unbiased estimator in terms of the Fisher information. In SfM/SLAM, the pose covariance of a bundle-adjustment (BA) solution relates to the inverse Hessian of the reprojection error. This is why CRBs have informed optimal design in pose-graph SLAM~\cite{chen2021}; vision methods plan viewpoints by maximizing Fisher information~\cite{zhang2019}. However, these analyses typically assume \emph{feature-based} measurements (e.g., 2D–3D correspondences). In contrast, Neural renderers give us a \emph{dense photometric} observation governed by a complex, differentiable image formation pipeline.

We address this gap by deriving a \emph{render-aware} CRB for pose on $\mathrm{SE}(3)$. We treat $I=R(\theta;x)$ as the observation model with fixed scene $\theta$ and pose $x\in \mathrm{SE}(3)$. Next, we can linearize image formation with respect to a tangent perturbation $\xi\in \mathfrak{se}(3)$, compute the per-pixel Jacobian $J=\partial R/\partial \xi$, and assemble a Fisher information matrix (FIM) $I(x)=J^\top \Sigma^{-1} J$. The bound $\mathrm{Cov}(\xi)\succeq I(x)^{-1}$ then quantifies the best-achievable pose accuracy.

Additionally, the eigenstructure of $I(x)$ exposes identifiability. High-texture, high-parallax regions yield information. Low-texture or symmetric content induces degeneracies (near-zero eigenvalues). Crucially, the formulation reduces to classical BA covariance in the pinhole/feature limit, providing continuity with established theory.

While our derivation begins with a single camera, we adopt the convention of treating each camera as an \emph{agent}. We show how this makes the formulation immediately extensible to multi-camera or cross-device settings, a useful downstream application. The method, in short, is to combine the Fisher information contributions from multiple agents, enabling efficient cooperative perception, fusion, and communication.

\textbf{Contributions.} 
(i) A general CRB for camera pose with differentiable renderers on $\mathrm{SE}(3)$;
(ii) practical autodiff recipes for per-ray Jacobians across NeRF/3DGS; 
(iii) links to BA/SLAM uncertainty and diagnostics for degeneracy; 
(iv) a compact protocol for empirical validation;
(v) a multi-agent extension for cooperative perception and fusion

\begin{figure}[h]
\centering
\includegraphics[width=1\linewidth]{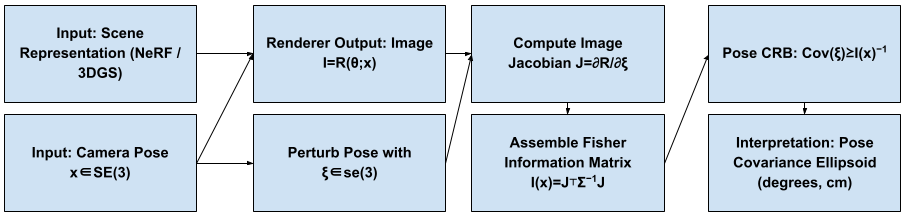}
\caption{Pipeline: fixed scene $\theta$ and pose $x$ $\rightarrow$ render $I$; autodiff gives $J=\partial R/\partial\xi$; FIM $J^\top\Sigma^{-1}J$; pose CRB $I(x)^{-1}$; interpret as ellipsoids in rotation/translation.}
\label{fig:pipeline}
\end{figure}

\section{Related Works}

\paragraph{Differentiable Rendering for Pose Estimation.}
Differentiable rendering can be used for camera pose estimation by enabling analysis-by-synthesis alignment. Neural rendering methods like NeRF provide dense and continuous scene representations that can produce photorealistic images given a camera pose. Following works (e.g. InstantNGP and 3D Gaussian Splatting) now provide fast differentiable image formation. Because of these advances, gradients of the rendering process can be used for pose optimization. For example, iNeRF (Inverting NeRF) demonstrated that a pretrained radiance field can be directly “inverted” to recover 6-DoF camera pose via gradient-based photometric alignment. Such works show how differentiable neural renderers, whether used post hoc for localization \cite{Lin2021IROS} or in-loop during mapping \cite{Lin2021ICCV}, can reliably estimate camera pose by minimizing pixel-wise reprojection error without explicit correspondences.

\paragraph{Uncertainty Quantification in Neural Rendering.}
Quantifying uncertainty in neural scene representations is a recent goal. Bayes’ Rays introduces a post-hoc Laplace approximation for NeRFs to estimate per-pixel confidence intervals \cite{Goli2024}. FisherRF leverages Fisher information to guide view selection and quantify parameter uncertainty \cite{Jiang2023}. Current directions are focused towards scene / model uncertainty. Our work is aimed towards uncertainty for camera poses given a fixed scene. By deriving a render-aware CRB on pose covariance, we provide a complementary, pose-centric analysis that captures geometric identifiability alongside model confidence.

\paragraph{Information-Theoretic Analyses of Camera Pose.}
Information theory provides a lens to evaluate and improve pose estimation. Chen et al. derive CRBs for pose-graph SLAM and propose optimal design metrics to distribute sensing effort \cite{chen2021}. Zhang and Scaramuzza extend this idea by introducing the Fisher Information Field for active visual localization \cite{zhang2019}. These approaches, however, assume feature-based measurements. In contrast, we treat a differentiable renderer as the observation model, yielding a dense photometric FIM for camera pose. By linearizing the full image formation process, our analysis bridges classical Fisher information methods with neural rendering, allowing us to quantify pose identifiability even without explicit correspondences.

\paragraph{Multi-Agent and Cooperative Perception.}
Multi-agent SLAM frameworks such as Kimera-Multi \cite{Tian2022} and COVINS \cite{Schmuck2021} demonstrate that sharing information across agents significantly improves localization accuracy and robustness. To build on top of this theme, we propose a principled method of fusing uncertainty by combining per-pixel Jacobians into a joint Fisher information matrix on a common reference frame. This yields a rigorous multi-agent pose CRB that ultimately aids cooperative view planning by communicating only the most informative observations.

\paragraph{Manifold and Statistical Estimation Foundations.}
Standard Lie-group state estimation and information theory are followed throughout our work. Barfoot’s text for $\mathrm{SE}(3)$ estimation~\cite{barfoot2017}, Sol\`a’s micro-Lie treatment and Jacobian calculus~\cite{sola2018}, and Riemannian optimization background~\cite{absil2008} justify local coordinates, reparameterization invariance, and reporting covariance in the tangent of $\mathrm{SE}(3)$.

\section{Methodology}
We define pose estimation as recovering a transformation $x\in \mathrm{SE}(3)$ from an image $I\in\mathbb{R}^{M}$ generated by a differentiable renderer $R$
\begin{equation}
I \;=\; R(\theta; x) + \eta,\qquad \eta\sim\mathcal{N}(0,\Sigma),
\label{eq:obs}
\end{equation}
with fixed scene parameters $\theta$ and pixel-noise covariance $\Sigma\in\mathbb{R}^{M\times M}$ (not necessarily diagonal). Let $\xi\in\mathfrak{se}(3)$ be a minimal twist so that the perturbed pose is $\exp(\xi)\,x$. Linearizing the image formation at $\xi=0$ gives
\begin{equation}
\begin{split}
R(\theta;\exp(\xi)\,x) \;\approx\; R(\theta;x) + J\,\xi,\\
J \;=\; \left.\frac{\partial R(\theta;\exp(\xi)\,x)}{\partial \xi}\right|_{\xi=0} \in \mathbb{R}^{M\times 6}.
\end{split}
\label{eq:lin}
\end{equation}

\subsection{Core Derivation}

\begin{theorem}[Render-aware Fisher information on $\mathrm{SE}(3)$]
\label{thm:fim}
Under the Gaussian model \eqref{eq:obs} and linearization \eqref{eq:lin}, the Fisher Information Matrix (FIM) for the local pose parameter $\xi$ is
\begin{equation}
\mathcal{I}(x) \;=\; J^\top \Sigma^{-1} J \;\in\; \mathbb{R}^{6\times 6},
\end{equation}
and the (unbiased) Cram\'er--Rao bound (CRB) on the local pose covariance is
\begin{equation}
\mathrm{Cov}(\hat{\xi}) \;\succeq\; \mathcal{I}(x)^{-1}.
\label{eq:crb}
\end{equation}
If $\mathcal{I}(x)$ is singular, interpret \eqref{eq:crb} using the Moore--Penrose pseudoinverse $\mathcal{I}(x)^+$.
\end{theorem}

\noindent\textit{Proof sketch.} For Gaussian $\eta$, $\log p(I\mid x) = -\tfrac{1}{2}(I-R(\theta;x))^\top \Sigma^{-1}(I-R(\theta;x))+\mathrm{const}$. Differentiating w.r.t.\ $\xi$ through \eqref{eq:lin} gives the score $\nabla_\xi \log p = J^\top \Sigma^{-1}(I-R(\theta;x))$ with mean $0$ and covariance $J^\top \Sigma^{-1} J$. The standard definition of the FIM as the covariance of the score gives $\mathcal{I}(x)$. The CRB follows. \qed

\paragraph{Reparameterization invariance.}
\begin{proposition}[Invariance to smooth minimal pose parametrization]
\label{prop:reparam}
Let $\phi:\mathbb{R}^6\!\to\!\mathbb{R}^6$ be a local diffeomorphism relating two minimal $\mathrm{SE}(3)$ coordinates $\xi$ and $\zeta=\phi(\xi)$. Then the information transforms as $\mathcal{I}_\zeta = (D\phi)^{-\top}\mathcal{I}_\xi (D\phi)^{-1}$ and the CRB \eqref{eq:crb} is invariant (up to the coordinate change). 
\end{proposition}
\noindent\textit{Remark.} The bound is thus well-defined on the manifold. We report rotation std in degrees and translation in scene units for interpretability.

\paragraph{Identifiability.}
\begin{lemma}[Local identifiability and degeneracy]
\label{lem:ident}
If the columns of $J$ span $\mathbb{R}^6$ on a set of nonzero measure pixels (equivalently, $\mathrm{rank}(J)=6$), then $\mathcal{I}(x)$ is full-rank and all pose directions are locally identifiable. If $J$ loses rank (e.g., constant-albedo planar wall, radial symmetry), $\mathcal{I}(x)$ becomes singular and the CRB diverges along the nullspace directions.
\end{lemma}

\paragraph{Classical BA as a special case.}
\begin{corollary}[Bundle adjustment (BA) limit]
\label{cor:ba}
If $R$ reduces to pinhole projection of known $3\mathrm{D}$ points $\{\mathbf{X}_k\}$ with per-point i.i.d.\ Gaussian noise $\sigma^2 I_2$, then stacking per-point reprojection Jacobians $J_k=\partial \pi(K[R|t]\mathbf{X}_k)/\partial \xi\in\mathbb{R}^{2\times 6}$ yields $J=\mathrm{blkrow}(J_k)$ and
\(
\mathcal{I}(x) = J^\top (\sigma^{-2} I) J,
\)
which equals the Gauss--Newton Hessian of reprojection BA; the CRB coincides with the BA covariance.
\end{corollary}

\subsection{Multi-Agent Extension}

This extension is critical for cooperative perception, where each camera contributes partial but complementary Fisher information.

\begin{figure}[h]
  \centering
  \includegraphics[width=\linewidth]{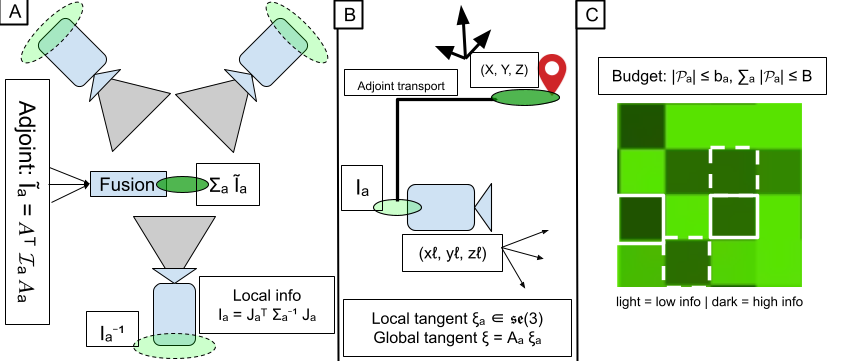}
  \caption{
  A) Multi-agent fusion of Fisher information. 
  B) Adjoint transport from local to global tangent. 
  C) Bandwidth-aware tile selection under budget constraints.}
  \label{fig:multiagent_extension}
\end{figure}

\textbf{Multi-agent FIM.}
For agents $a\!=\!1{:}A$ with image Jacobians $J_a$ and noise $\Sigma_a$, the per-agent information in the agent’s local tangent is $\mathcal{I}_a = J_a^\top \Sigma_a^{-1} J_a$.
To fuse in a global pose tangent (about $x$), we transport via the $SE(3)$ adjoint: $\tilde{\mathcal{I}}_a = A_a^\top \mathcal{I}_a A_a$, where $A_a \!=\! \operatorname{Ad}_{g_a^{-1}}$ maps the agent’s local perturbations to the global frame (here $g_a$ is the relative transform between frames, Fig.~\ref{fig:multiagent_extension}B).
A concrete form is
\[
\operatorname{Ad}_{g} \;=\;
\begin{bmatrix}
R & [t]_\times R\\
0 & R
\end{bmatrix},
\quad g=\begin{bmatrix}R & t\\ 0 & 1\end{bmatrix}\in SE(3),
\]
with $[t]_\times$ the skew-symmetric matrix of $t$.
Under conditional independence of pixel noise given $(\theta,x)$, the joint information is
\[
\mathcal{I}_{\text{joint}}(x)=\sum_{a=1}^A \tilde{\mathcal{I}}_a.
\]
In an information-filter view, communicating $\tilde{\mathcal{I}}_a$ (or its Cholesky/eigen-sketch) yields consistent fusion under bandwidth limits (Fig.~\ref{fig:multiagent_extension}A).

\textbf{Bandwidth-aware agent/tile selection.}
Partition each image into tiles $\{\mathcal{T}_{a,t}\}$ with tile-level Fisher blocks $\tilde{\mathcal{I}}_{a,t}$ (Fig.~\ref{fig:multiagent_extension}C).
Given per-agent budgets $b_a$ and a global budget $B$, select $\mathcal{P}_a \subseteq \{\mathcal{T}_{a,t}\}$ to maximize
\[
f\!\Big(\mathcal{I}_0 + \sum_{a}\sum_{t\in\mathcal{P}_a} \tilde{\mathcal{I}}_{a,t}\Big),\quad
\text{s.t. }\sum_a |\mathcal{P}_a|\le B,\;\;|\mathcal{P}_a|\le b_a.
\]
We use $f\in\{\log\det(\cdot),\,\mathrm{tr}(\cdot),\,\lambda_{\min}(\cdot)\}$.
$\log\det$ is monotone submodular (greedy gives a $(1\!-\!1/e)$ approximation under cardinality/partition constraints), $\mathrm{tr}$ is modular (greedy is optimal), while $\lambda_{\min}$ is not submodular (greedy is a heuristic).
In practice we add a small ridge $\epsilon I$ for numerical stability when computing $f$.

\subsection{Computing $J$ in practice (autodiff and VJPs)}

\begin{algorithm}[h]
\caption{CRB via implicit Jacobians (JVPs)}
\label{alg:crb_jvp}
\begin{algorithmic}[1]
\REQUIRE Renderer $R(\theta;x)$; pose $x$; noise model $\Sigma$ (apply $w\leftarrow\Sigma^{-1}v$); pixel subset $\mathcal{P}\subset\{1,\dots,M\}$
\STATE Define $f(\xi)=R(\theta;\exp(\xi)x)$ and evaluate at $\xi=0$
\FOR{$j=1$ to $6$}
  \STATE $q_j \leftarrow \mathrm{JVP}_f(e_j)$ restrict to pixels $\mathcal{P}$  // column $j$ of $J$
  \STATE $u_j \leftarrow \Sigma^{-1} q_j$  // elementwise if $\Sigma$ is (block-)diagonal
\ENDFOR
\STATE $\mathcal{I}_{ij} \leftarrow \langle q_i, u_j\rangle_{\mathcal{P}} \;\;(i,j=1..6)$ \quad // $\mathcal{I}=J^\top\Sigma^{-1}J$
\STATE \textbf{return} $\widehat{\mathcal{I}}(x)$ and
\[
\widehat{C}=\begin{cases}
\widehat{\mathcal{I}}^{-1}, & \text{if $\widehat{\mathcal{I}}$ is PD},\\[4pt]
\widehat{\mathcal{I}}^{+}, & \text{otherwise (Moore--Penrose, optional ridge $\epsilon I$)}.
\end{cases}
\]
\end{algorithmic}
\end{algorithm}

Forming $J$ directly by per-pixel gradients is memory-intensive. So, we instead exploit vector-Jacobian products (VJPs). For any vector $v\in\mathbb{R}^M$, autodiff gives $J^\top v$ without materializing $J$. This is adequate for assembling $\mathcal{I}(x)=J^\top \Sigma^{-1} J$ by applying $\Sigma^{-1}$ to columns of $J$ implicitly. For diagonal (or block-diagonal) $\Sigma$, $\Sigma^{-1}$ is cheap. Pixel subsampling and tiling can further reduce cost.

\paragraph{Complexity and scalability.}
Let $|\mathcal{P}|$ be the number of pixels that are sampled. Forming $\mathcal{I}(x)$ requires $6$ columns $J e_j$ and their weighted inner products: $O(6\,|\mathcal{P}|)$ renderer VJPs plus cheap reductions for diagonal $\Sigma$. With $|\mathcal{P}|=s M$ (subsampling rate $s\in(0,1]$), cost scales linearly in $sM$. Tiling reduces memory, and blockwise accumulation avoids storing $J$. This approach would be practical for $512^2$ images on modern GPUs.

\subsection{Modeling assumptions and robustness}
\textbf{Noise.} The derivation holds for general (possibly correlated) noise $\Sigma$. In practice, per-pixel variances $\hat{\Sigma}=\mathrm{diag}(\hat{\sigma}_i^2)$ can be estimated from residuals. Larger noise weakens the bound.  
\textbf{Photometry.} Illumination drift or tone-mapping mismatches bias $J$ and the FIM. Normalization, learned $\hat{\Sigma}$, or restricting to gradient-rich pixels can mitigate this.  
\textbf{Bias.} The CRB applies to unbiased estimators. At high SNR, MLEs approach the bound. Biased extensions (e.g., van Trees) are possible but omitted here.

\paragraph{Interpretation and reporting.}
$\sqrt{\mathrm{diag}(\mathcal{I}(x)^{-1})}$ is reported as $1\sigma$ pose bounds (rotation in degrees, translation in scene units). Eigenvalues of $\mathcal{I}(x)$ highlight ill-conditioning.

\paragraph{Practitioner recipe.}
(i) Freeze $\theta$; (ii) treat pose as a 6D input; (iii) autodiff on a pixel subset to compute $J e_j$; (iv) weight by $\Sigma^{-1}$; (v) assemble $\mathcal{I}(x)$ and invert (or pseudoinvert); (vi) inspect eigenstructure.

\section{Experiments}
Code released at \url{https://github.com/ArunMut/Multi-Agent-Pose-Uncertainty}

We validate the render-aware CRB on Instant-NGP~\cite{mueller2022} and 3D Gaussian Splatting~\cite{kerbl2023} across LLFF \cite{mildenhall2019} (texture-rich) and Tanks \& Temples \cite{knapitsch2017} (often low-texture). For each scene, we compute the pose FIM from per-pixel Jacobians and compare the resulting CRB to (i) empirical pose errors from small perturb-and-align trials (iNeRF-style~\cite{Lin2021IROS}) and (ii) pose covariances from bundle adjustment (BA) when feature tracks are available.

Starting from a known pose $x$, we render $I$, perturb $x$ by a small random $\Delta x$, and realign by gradient descent to obtain $\hat{x}$. Across many trials, rotation/translation RMSE closely matches the CRB: high-texture scenes yield sub-degree and $\sim$centimeter bounds, while low-texture scenes show multi-degree and decimeter-scale bounds (Table~\ref{tab:results}). When keypoints are available, BA covariances (from the Hessian inverse) also agree with our CRB in well-conditioned views, with differences of only a few percent. In degenerate cases such as a planar white wall, the FIM has near-zero eigenvalues along translation parallel to the wall and rotation about the optical axis, so the pseudoinverse $I(x)^{+}$ produces very large variances in those modes, consistent with BA and geometric intuition.

\begin{table}[h]\small
\centering
\begin{tabular}{lcc}
\toprule
\textbf{Scenario} & \textbf{Rot. error (deg)} & \textbf{Trans. error (cm)} \\
\midrule
High-texture (CRB)       & 0.4 & 1.3 \\
High-texture (Empirical) & 0.5 & 1.5 \\
High-texture (BA Cov)    & 0.2 & 0.9 \\
\midrule
Low-texture (CRB)        & 5.1 & 21 \\
Low-texture (Empirical)  & 5.5 & 23 \\
Low-texture (BA Cov)     & 4.9 & 19 \\
\bottomrule
\end{tabular}
\caption{CRB vs. empirical pose error and BA covariance. Texture-rich views are tightly constrained; low-texture views are ill-conditioned. The CRB tracks both empirical and BA uncertainties.}
\label{tab:results}
\end{table}

We further evaluate two aspects of the bound: calibration and cooperative gains.

\begin{figure}[h]
    \centering
    \includegraphics[width=\linewidth]{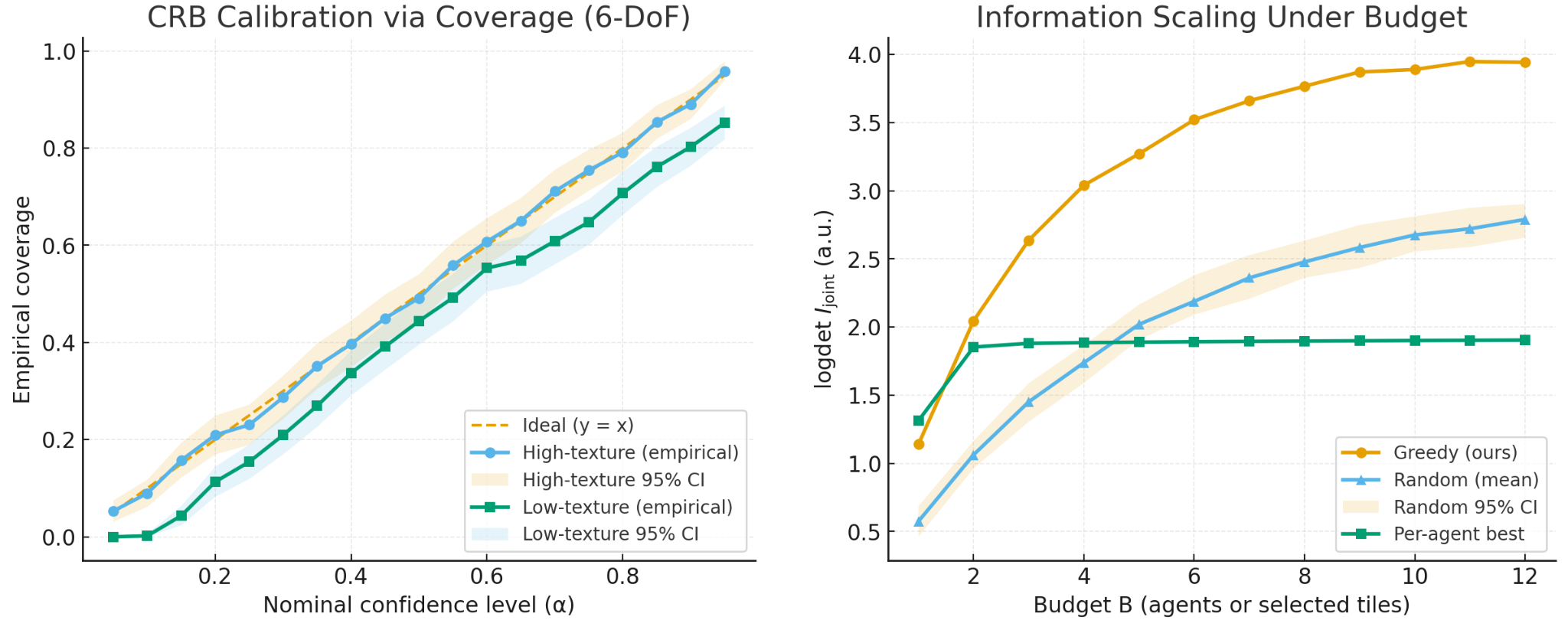}
    \caption{CRB calibration and cooperative gains. 
    \textbf{Left:} Coverage vs.\ nominal confidence shows calibration in high-texture scenes and under-coverage in low-texture ones. 
    \textbf{Right:} log-det information grows submodularly with budget; greedy selection outperforms random and per-agent baselines.}
    \label{fig:crb_calib_scaling}
\end{figure}

In high-texture scenes, empirical coverage aligns closely with nominal confidence, while greedy selection yields nearly twice the information of random baselines under the same budget (Fig.~\ref{fig:crb_calib_scaling}). 
These results suggest that the CRB can serve as both a diagnostic tool for view quality and a principled signal for multi-agent view planning.

\section{Conclusion}
We derived a render-aware Fisher information and Cramér--Rao bound on $\mathrm{SE}(3)$, showing how texture and geometry govern pose identifiability. The bound reduces to bundle adjustment in classical cases, matches empirical errors, and extends naturally to multi-agent settings via Fisher information fusion. Future work will address dynamic scenes and use the bound for view planning and adaptive rendering.

{
    \small
    \bibliographystyle{ieeenat_fullname}
    \nocite{*}
    \bibliography{main}

\begin{thebibliography}{21}
\providecommand{\natexlab}[1]{#1}
\providecommand{\url}[1]{\texttt{#1}}
\expandafter\ifx\csname urlstyle\endcsname\relax
  \providecommand{\doi}[1]{doi: #1}\else
  \providecommand{\doi}{doi: \begingroup \urlstyle{rm}\Url}\fi

\bibitem[Absil et~al.(2008)Absil, Mahony, and Sepulchre]{absil2008}
P.-A. Absil, Robert Mahony, and Rodolphe Sepulchre.
\newblock \emph{Optimization Algorithms on Matrix Manifolds}.
\newblock Princeton University Press, 2008.

\bibitem[Alismail et~al.(2016)Alismail, Browning, and Lucey]{alismail2016photometricBA}
Hatem Alismail, Brett Browning, and Simon Lucey.
\newblock Photometric bundle adjustment for vision-based slam.
\newblock \emph{arXiv preprint arXiv:1608.02026}, 2016.

\bibitem[Baker and Matthews(2004)]{baker2004lk20years}
Simon Baker and Iain Matthews.
\newblock Lucas--kanade 20 years on: A unifying framework.
\newblock \emph{International Journal of Computer Vision}, 56\penalty0 (3):\penalty0 221--255, 2004.

\bibitem[Barfoot(2017)]{barfoot2017}
Timothy~D. Barfoot.
\newblock \emph{State Estimation for Robotics}.
\newblock Cambridge University Press, 2017.

\bibitem[Chen et~al.(2021)Chen, Huang, Zhao, and Dissanayake]{chen2021}
Yongbo Chen, Shoudong Huang, Liang Zhao, and Gamini Dissanayake.
\newblock Cram\'er--rao bounds and optimal design metrics for pose-graph slam.
\newblock \emph{IEEE Transactions on Robotics}, 37\penalty0 (2):\penalty0 627--641, 2021.

\bibitem[Delaunoy and Pollefeys(2014)]{delaunoy2014photometricBA}
Amaury Delaunoy and Marc Pollefeys.
\newblock Photometric bundle adjustment for dense multi-view 3d modeling.
\newblock In \emph{Proc. IEEE/CVF Conf. Computer Vision and Pattern Recognition (CVPR)}, pages 1486--1493, 2014.

\bibitem[Engel et~al.(2014)Engel, Sch{\"o}ps, and Cremers]{engel2014lsdslam}
Jakob Engel, Thomas Sch{\"o}ps, and Daniel Cremers.
\newblock {LSD-SLAM}: Large-scale direct monocular slam.
\newblock In \emph{Proc. European Conference on Computer Vision (ECCV)}, pages 834--849. Springer, 2014.

\bibitem[Engel et~al.(2018)Engel, Koltun, and Cremers]{engel2018dso}
Jakob Engel, Vladlen Koltun, and Daniel Cremers.
\newblock Direct sparse odometry.
\newblock \emph{IEEE Transactions on Pattern Analysis and Machine Intelligence}, 40\penalty0 (3):\penalty0 611--625, 2018.

\bibitem[Goli et~al.(2024)Goli, Reading, Sell{\'a}n, Jacobson, and Tagliasacchi]{Goli2024}
Lily Goli, Cody Reading, Silvia Sell{\'a}n, Alec Jacobson, and Andrea Tagliasacchi.
\newblock Bayes' rays: Uncertainty quantification for neural radiance fields.
\newblock In \emph{IEEE/CVF Conference on Computer Vision and Pattern Recognition (CVPR)}, 2024.

\bibitem[Jiang et~al.(2024)Jiang, Lei, and Daniilidis]{Jiang2023}
Wen Jiang, Boshu Lei, and Kostas Daniilidis.
\newblock Fisherrf: Active view selection and mapping with radiance fields using fisher information.
\newblock 2024.
\newblock Extended from arXiv:2311.17874.

\bibitem[Kerbl et~al.(2023)Kerbl, Kopanas, Leimk{\"u}hler, and Drettakis]{kerbl2023}
Bernhard Kerbl, Georgios Kopanas, Thomas Leimk{\"u}hler, and George Drettakis.
\newblock 3d gaussian splatting for real-time radiance field rendering.
\newblock \emph{ACM Transactions on Graphics (SIGGRAPH)}, 42\penalty0 (4), 2023.

\bibitem[Knapitsch et~al.(2017)Knapitsch, Park, Zhou, and Koltun]{knapitsch2017}
Arno Knapitsch, Jaesik Park, Qian-Yi Zhou, and Vladlen Koltun.
\newblock Tanks and temples: Benchmarking large-scale scene reconstruction.
\newblock \emph{ACM Transactions on Graphics (ToG)}, 36\penalty0 (4):\penalty0 1--13, 2017.

\bibitem[Lin et~al.(2021{\natexlab{a}})Lin, Ma, Torralba, and Lucey]{Lin2021ICCV}
Chen-Hsuan Lin, Wei-Chiu Ma, Antonio Torralba, and Simon Lucey.
\newblock {BARF}: Bundle-adjusting neural radiance fields.
\newblock In \emph{IEEE/CVF International Conference on Computer Vision (ICCV)}, pages 5741--5751, 2021{\natexlab{a}}.

\bibitem[Lin et~al.(2021{\natexlab{b}})Lin, Florence, Barron, Rodriguez, Isola, and Lin]{Lin2021IROS}
Yen-Chen Lin, Pete Florence, Jonathan~T. Barron, Alberto Rodriguez, Phillip Isola, and Tsung-Yi Lin.
\newblock {iNeRF}: Inverting neural radiance fields for pose estimation.
\newblock In \emph{IEEE/RSJ International Conference on Intelligent Robots and Systems (IROS)}, pages 1323--1330, 2021{\natexlab{b}}.

\bibitem[Mildenhall et~al.(2019)Mildenhall, Srinivasan, Ortiz-Cayon, Kalantari, Ramamoorthi, Ng, and Kar]{mildenhall2019}
Ben Mildenhall, Pratul~P Srinivasan, Rodrigo Ortiz-Cayon, Nima~Khademi Kalantari, Ravi Ramamoorthi, Ren Ng, and Abhishek Kar.
\newblock Local light field fusion: Practical view synthesis with prescriptive sampling guidelines.
\newblock \emph{ACM Transactions on Graphics (ToG)}, 38\penalty0 (4):\penalty0 1--14, 2019.

\bibitem[Mildenhall et~al.(2020)Mildenhall, Srinivasan, Tancik, Barron, Ramamoorthi, and Ng]{mildenhall2020}
Ben Mildenhall, Pratul~P. Srinivasan, Matthew Tancik, Jonathan~T. Barron, Ravi Ramamoorthi, and Ren Ng.
\newblock {NeRF}: Representing scenes as neural radiance fields for view synthesis.
\newblock In \emph{European Conference on Computer Vision (ECCV)}, 2020.

\bibitem[M{\"u}ller et~al.(2022)M{\"u}ller, Evans, Schied, and Keller]{mueller2022}
Thomas M{\"u}ller, Alex Evans, Christoph Schied, and Alexander Keller.
\newblock Instant neural graphics primitives with a multiresolution hash encoding.
\newblock \emph{ACM Transactions on Graphics (SIGGRAPH)}, 41\penalty0 (4):\penalty0 102:1--102:15, 2022.

\bibitem[Schmuck et~al.(2021)Schmuck, Ziegler, Karrer, Perraudin, and Chli]{Schmuck2021}
Patrik Schmuck, Thomas Ziegler, Marco Karrer, Jonathan Perraudin, and Margarita Chli.
\newblock {COVINS}: Visual-inertial slam for centralized collaboration.
\newblock In \emph{IEEE International Symposium on Mixed and Augmented Reality (ISMAR) -- Adjunct}, 2021.

\bibitem[Sol{\`a} et~al.(2018)Sol{\`a}, Deray, and Atchuthan]{sola2018}
Joan Sol{\`a}, Jeremie Deray, and Dinesh Atchuthan.
\newblock A micro lie theory for state estimation in robotics.
\newblock \emph{arXiv preprint arXiv:1812.01537}, 2018.

\bibitem[Tian et~al.(2022)Tian, Chang, Arias, Nieto-Granda, How, and Carlone]{Tian2022}
Yulun Tian, Yun Chang, Fernando~Herrera Arias, Carlos Nieto-Granda, Jonathan~P. How, and Luca Carlone.
\newblock Kimera-multi: Robust, distributed, dense metric-semantic slam for multi-robot systems.
\newblock \emph{IEEE Transactions on Robotics}, 38\penalty0 (4):\penalty0 2022--2038, 2022.

\bibitem[Zhang and Scaramuzza(2019)]{zhang2019}
Zichao Zhang and Davide Scaramuzza.
\newblock Beyond point clouds: Fisher information field for active visual localization.
\newblock In \emph{IEEE International Conference on Robotics and Automation (ICRA)}, pages 5984--5990, 2019.

\end{thebibliography}
}


\end{document}